\documentclass[letterpaper, 10 pt, conference]{ieeeconf}  

\IEEEoverridecommandlockouts                              

\usepackage{graphicx} 
\usepackage{amsmath} 
\usepackage{amsfonts}
\usepackage{bm}
\usepackage{cite}
\usepackage{mathtools}
\usepackage[ruled,vlined]{algorithm2e}
\usepackage{booktabs}
\usepackage{multirow}
\usepackage{hyperref}
\usepackage{amssymb}

\hypersetup{
hidelinks,
colorlinks=false,
linkcolor=blue,
citecolor=black
}

\title{\LARGE \bf
Disturbance-Aware Dynamical Trajectory Planning for Air-Land Bimodal Vehicles
}

 \author{Shaoting Liu$^{1,2}$, Wenshuai Yu$^{2}$, Bo Zhang$^{2}$, Shoubin Chen$^{1}$, Fei Ma$^{1}$, Zhou Liu$^{1,*}$, Qingquan Li$^{2}$       
\thanks{* denotes corresponding author.}
\thanks{$^{1}$Shaoting Liu, Shoubin Chen, Fei Ma and Zhou Liu are with the Guangdong Laboratory of Artificial Intelligence and Digital Economy (Shenzhen), Shenzhen 518107, China.}%
\thanks{$^{2}$Shaoting Liu, Wenshuai Yu, Bo Zhang, and Qingquan Li are with Shenzhen University, Shenzhen 518060, China.}%
}

\begin{document}
\bibliographystyle{unsrt}

\maketitle
\thispagestyle{empty}
\pagestyle{empty}

\begin{abstract}


Air–land bimodal vehicles provide a promising solution for navigating complex environments by combining the flexibility of aerial locomotion with the energy efficiency of ground mobility. However, planning dynamically feasible, smooth, collision-free, and energy-efficient trajectories remains challenging due to two key factors: 1) unknown dynamic disturbances in both aerial and terrestrial domains, and 2) the inherent complexity of managing bimodal dynamics with distinct constraint characteristics. This paper proposes a disturbance-aware motion-planning framework that addresses this challenge through real-time disturbance estimation and adaptive trajectory generation. The framework comprises two key components: 1) a disturbance-adaptive safety boundary adjustment mechanism that dynamically determines the feasible region of dynamic constraints for both air and land modes based on estimated disturbances via a disturbance observer, and 2) a constraint-adaptive bimodal motion planner that integrates disturbance-aware path searching to guide trajectories toward regions with reduced disturbances and B-spline-based trajectory optimization to refine trajectories within the established feasible constraint boundaries. Experimental validation on a self-developed air-land bimodal vehicle demonstrates substantial performance improvements across three representative disturbance scenarios, achieving an average 33.9\% reduction in trajectory tracking error while still maintaining superior time-energy trade-offs compared to existing methods.

\end{abstract}

\section{Introduction}
Unmanned Aerial Vehicles (UAVs), particularly quadrotors, are widely employed for their exceptional maneuverability, hovering capability, and ability to navigate complex terrains, making them ideal for search and rescue, agriculture, and delivery \cite{tian2025uavs}. However, a major limitation is their high energy consumption, since substantial power is required to counteract gravity \cite{lee2020fail}, thereby restricting their endurance. Typically, UAVs exhibit operational flight times ranging from 5 to 30 minutes \cite{dai2019analytical}.

In contrast, Unmanned Ground Vehicles (UGVs) provide higher energy efficiency, greater payload capacity, and quieter operation, making them particularly well-suited for logistics and extended-duration missions in structured environments. However, they face challenges in terrain adaptability and are generally less effective than UAVs in traversing complex obstacles \cite{zheng2024capsulebot}. This limitation significantly reduces their effectiveness in unstructured or rugged terrains \cite{araki2017multi}.

\begin{figure}[thpb]
    \centering
    \includegraphics[width=1\linewidth]{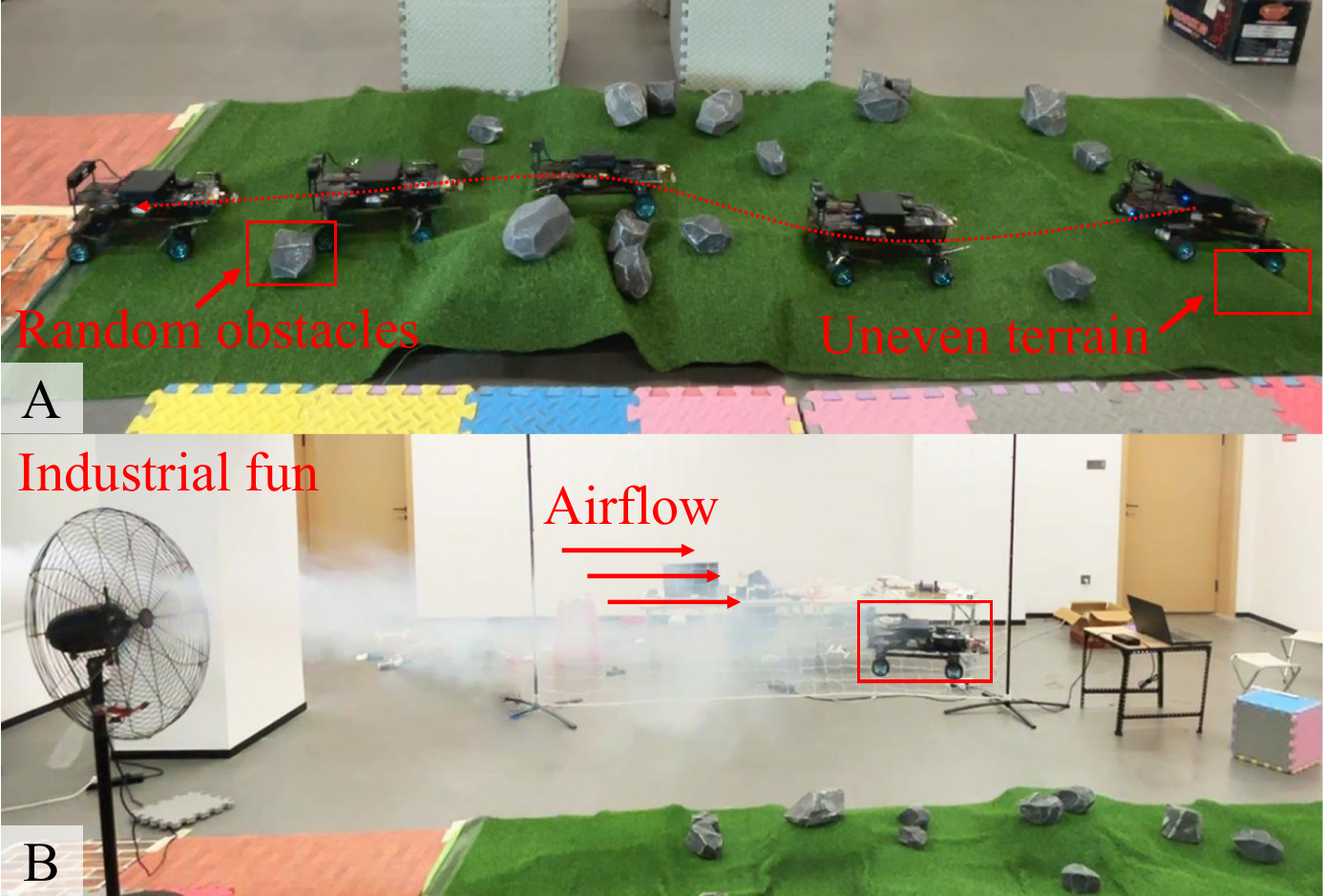}
    \caption{Snapshots of the air-land bimodal vehicle navigating in environments with different types of disturbances. (A) Terrestrial disturbance scenario: the vehicle traverses an uneven terrain with randomly placed obstacles. (B) Aerial disturbance scenario: an industrial fan generates strong airflow, visualized by smoke, to simulate wind disturbances during flight.}
    \label{fig:airflow}
\end{figure}

To address the respective limitations of UAVs and UGVs, researchers have proposed various air-land bimodal vehicles that combine the aerial agility of UAVs with the ground efficiency of UGVs. Several notable design paradigms have emerged in the field of air–land bimodal vehicles. For instance, Yamada et al. \cite{yamada2017development} introduced a passive‑wheel bimodal vehicle, which incorporates a quadrotor for aerial flight and passive wheels for ground locomotion. Similarly, Zhang et al. \cite{zhang2022autonomous,zhang2023model} proposed a passive‑wheel terrestrial–aerial bimodal vehicle together with a navigation framework that enables autonomous flight. Tan et al. \cite{tan2021multimodal} introduced a fly–drive vehicle integrating a hexacopter flight system with a four‑wheel ground mobility system. There are also deformable bimodal vehicle designs \cite{david2021design,morton2017small}, which can fold into a vehicle structure in wheeled land mode. 

Due to the complexity of hybrid platforms, efficient motion planning is essential for fully exploiting their multimodal capabilities. Motion planning for UAVs has been extensively studied, providing a solid foundation for subsequent developments in bimodal platforms. Zhou et al.\cite{zhou2019robust} proposed a hierarchical motion planning framework for quadrotors that integrates a kinodynamic A* path-searching front end with a B-spline optimization back end, enabling robust and efficient flight in cluttered environments. Zhou et al.\cite{zhou2020ego} introduced an ESDF-free, gradient-based planning framework that reduces redundant computations by formulating collision costs through comparisons with a guiding path, thereby enabling efficient trajectory optimization with improved robustness. Sonny et al.\cite{sonny2023q} developed a Q-learning-based UAV path planning method that integrates a Shortest Distance Prioritization policy with a grid-graph formulation, enabling efficient navigation with both static and dynamic obstacle avoidance and achieving superior performance in path length and learning time compared with A*, Dijkstra. Building on advances in single-modal systems, researchers have extended motion planning to air–land bimodal vehicles. For instance, Fan et al.\cite{fan2019autonomous} proposed an autonomous navigation framework for hybrid ground–aerial vehicles that accounts for the additional energy consumption of aerial movements. Zhang et al.\cite{zhang2022autonomous} proposed a terrestrial–aerial bimodal planner and controller that also considers nonholonomic constraints of ground motion. The incorporation of curvature constraints ensures that the ground trajectory remains consistent with vehicle dynamics. Wang et al.\cite{wang2024agrnav} proposed AGRNav, a navigation framework for air–ground robots that leverages a lightweight semantic scene completion network for obstacle prediction and a hierarchical planner for energy-efficient path search, achieving superior performance in experiments. However, these frameworks generally assume disturbance-free conditions and have not been validated in environments with significant disturbances, which limits their practical applicability in real-world scenarios.

In control systems, disturbance observers have been widely employed to estimate and compensate for unknown disturbances, thereby improving controller tracking performance \cite{zhang2018ude, jia2022agile, tal2020accurate}. However, how to effectively leverage disturbance information to enhance the planning performance of air–land bimodal vehicles in disturbed environments remains an open problem. Motivated by the proven advantages of disturbance observers in control and the limitations of existing planners in disturbance handling, this paper proposes a disturbance-aware planning framework that incorporates disturbance estimation into the motion planner of air–land bimodal vehicles to improve robustness against environmental disturbances. The primary contributions of this paper are summarized as follows:

\begin{enumerate}

\item A disturbance-adaptive safety boundary adjustment mechanism is proposed that employs a disturbance observer to estimate environmental disturbances and dynamically determines the feasible region of dynamic constraints for both air and land modes.

\item A constraint-adaptive motion planner is proposed that comprises two integrated modules: a disturbance-aware path-searching front end for generating feasible initial trajectories under disturbances, and a B-spline-based trajectory optimization back end for refining these trajectories. Operating within the established bimodal dynamic constraint boundaries, the planner generates dynamically feasible, collision-free, smooth, and energy-efficient trajectories.

\item To the best of our knowledge, this is the first work to realize autonomous navigation for air–land bimodal vehicles in the presence of environmental disturbances. Simulations and real-world experiments across three representative disturbance environments were conducted to validate its effectiveness and robustness. Results show an average 33.9\% reduction in trajectory tracking error while maintaining superior time–energy trade-offs compared with existing methods under highly disturbed conditions.

\end{enumerate}

\section{Problem Formulation}

To describe the movement mechanism of the air–land bimodal vehicle, we establish the translational dynamics models for both the quadrotor flight mode and the wheeled land mode.

\begin{figure}[ht]
    \centering
    \includegraphics[width=\linewidth]{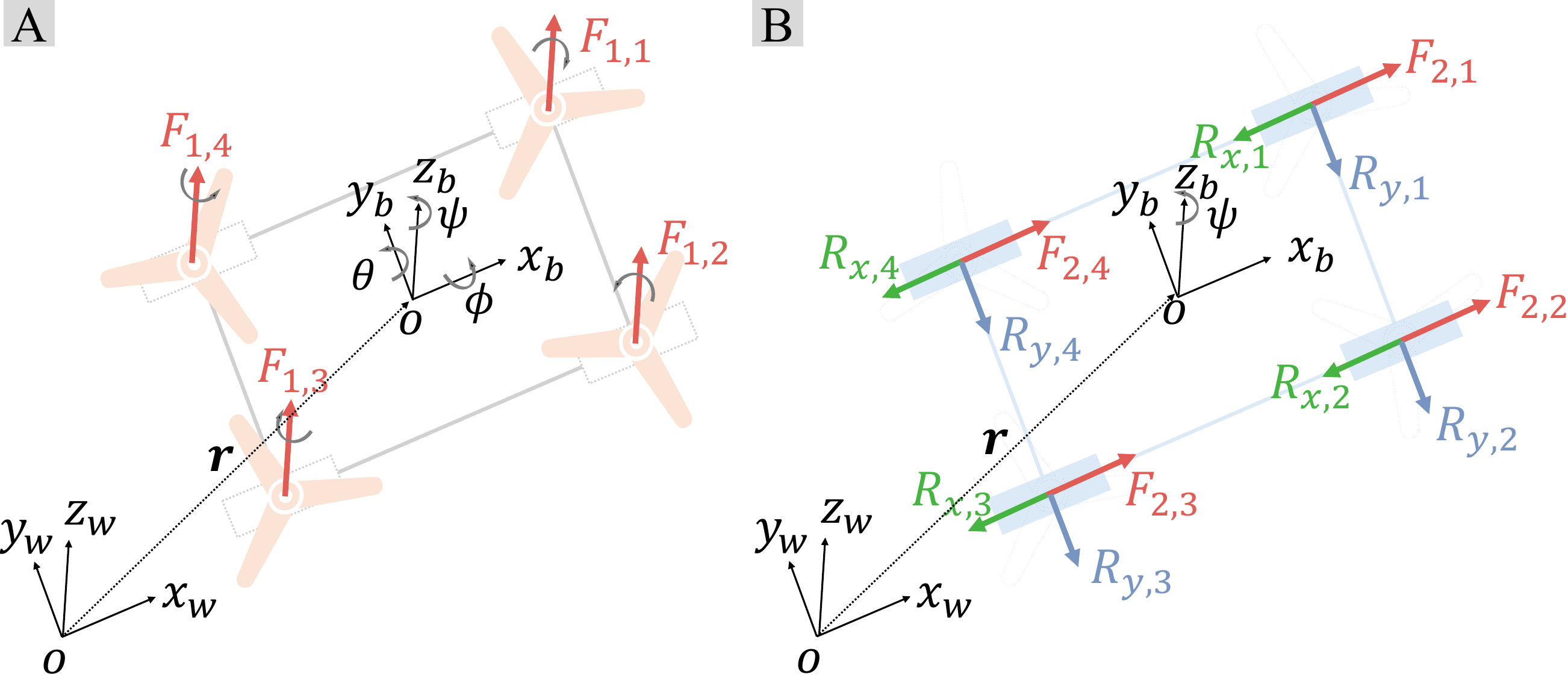}
    \caption{(A) Translational dynamics model of the quadrotor flight mode. (B) Translational dynamics model of the wheeled land mode.}
    \label{fig:dynamics}
\end{figure}

\subsection{Translational Dynamics of the Quadrotor Flight Mode}

In flight mode, the bimodal vehicle is modeled as a quadrotor (Fig.~\ref{fig:dynamics}A). The position of the vehicle in the world frame $\{x_w,y_w,z_w\}$ is represented by $\mathbf{r} = [x,y,z]^T \in \mathbb{R}^3$, and the rotation from body frame $\{x_b,y_b,z_b\}$ to world frame is defined by three Euler angles $\mathbf {\Theta} = [\phi,\theta,\psi]^T \in \mathbb{R}^3$, including roll, pitch and yaw, respectively. Its translational dynamics can be expressed as \cite{liu2023simultaneous,8352057,6844882}:
\begin{equation}
m\ddot{\mathbf{r}} = F_1{\mathbf{R}}_3(\mathbf{\Theta}) - \mathbf{d}_\text{air} - m\mathbf{g},
\label{eq:trans}
\end{equation}
where $m$ is the vehicle mass, $F_1=\sum_{i=1}^{4}F_{1,i}$ is the total thrust, ${\mathbf{R_3}}({\mathbf{\Theta}})$ is the Euler angle vector as
\begin{equation}
{\mathbf{R_3}}({\mathbf{\Theta}}) = 
\begin{bmatrix}
\cos\phi \sin\theta \cos\psi  + \sin\phi \sin\psi \\
\cos\phi \sin\theta \sin\psi  - \sin\phi \cos\psi \\
\cos\theta \cos\phi
\end{bmatrix},
\end{equation}
$\mathbf{d}_\text{air}=[d_{air,1},d_{air,2},d_{air,3}]^T \in \mathbb{R}^3$ is the external disturbance vector, and $\mathbf{g}=[0,0,g]^T \in \mathbb{R}^3$ denotes the gravity acceleration vector.

\subsection{Translational Dynamics of the Wheeled Land Mode}

In land mode, the vehicle is modeled as a four-wheel differential-drive system (Fig.~\ref{fig:dynamics}B). Since the vehicle is assumed to travel on a flat plane, vertical motion is negligible. Accordingly, its position in the world frame is denoted by $\mathbf{r} = [x, y, 0]^T \in \mathbb{R}^3$.  

A commonly adopted modeling simplification is that the two wheels on each side generate equal driving forces \cite{8458441,b2d27ee96d6c4f4c9683f12fec1a7583,wu2013differential}, i.e., $F_{2,1}=F_{2,4}$ and $F_{2,2}=F_{2,3}$. Within this modeling framework, the translational dynamics are given by:
\begin{equation}
{\mathbf{M\ddot r}} = \mathbf{C}(\psi){\mathbf{F}} - \mathbf{H}(\psi){\mathbf{d}_\text{land}},
\label{eq:ground-dynamic-uni}
\end{equation}
where
\begin{equation}
\mathbf{M} = 
\begin{bmatrix}
m & 0 & 0 \\
0 & m & 0 \\
0 & 0 & 1
\end{bmatrix}, \quad
\mathbf{C}(\psi) = 
\begin{bmatrix}
2\cos\psi & 2\cos\psi \\
2\sin\psi & 2\sin\psi \\
0 & 0
\end{bmatrix},
\end{equation}
$\mathbf{F} = [F_{2,1}, F_{2,2}]^T\in\mathbb{R}^2$ denotes the vector of driving forces, 
\begin{equation}
\mathbf{H}(\psi) =
\begin{bmatrix}
\cos\psi & -\sin\psi & 0 \\
\sin\psi & \cos\psi & 0 \\
0 & 0 & 1
\end{bmatrix},
\end{equation}
and $\mathbf{d}_\text{land} = [R_x, R_y, 0]^T\in\mathbb{R}^3$ denotes the disturbance vector, where $R_x = \sum_{i=1}^4 R_{x,i}$ and $R_y = \sum_{i=1}^4 R_{y,i}$ denote the longitudinal and lateral resistive forces, respectively.

\subsection{Research Objective}
Given the dynamics of both the quadrotor flight mode (\ref{eq:trans}) and the wheeled land mode (\ref{eq:ground-dynamic-uni}), along with the vehicle’s current and target states, an obstacle set, and the actuation limits—namely the maximum thrust (in flight mode) and maximum driving force (in land mode) that the air–land bimodal vehicle can generate along the body-frame axes \{$x_b, y_b, z_b$\}, this paper aims to develop a disturbance-aware motion-planning framework that generates a feasible trajectory from the current state to the target state. The resulting trajectory ensures robustness to disturbances, compliance with air–land bimodal dynamics, energy efficiency, smoothness, and obstacle avoidance.


\section{Disturbance-Adaptive Safety Boundary Adjustment}
\label{sect:Disturbance-Adaptive Safety Boundary Adjustment}
\begin{figure}[htbp]
    \centering
    \includegraphics[width=\linewidth]{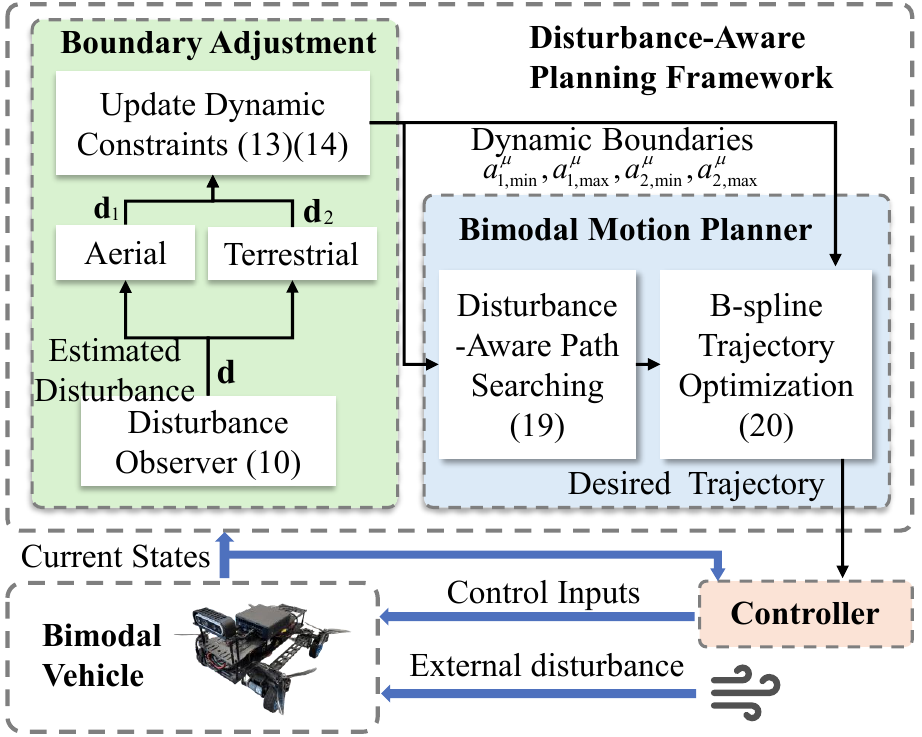}
    \caption{Overview of the proposed disturbance-aware planning framework.}
    \label{fig:planning-framework}
\end{figure}

To achieve the aforementioned research objective, we propose a disturbance-aware planning framework, as illustrated in Fig.~\ref{fig:planning-framework}. The framework comprises two key components: a disturbance-adaptive safety-boundary adjustment mechanism (shown in the green box) and a constraint-adaptive bimodal motion planner (shown in the blue box). This section first elaborates on the disturbance-adaptive safety-boundary adjustment mechanism in detail.

\subsection{Disturbance Observer}
\label{sec:Disturbance Observer}

The disturbance observer (DOB) is designed to estimate the external disturbances acting on the dynamic system. To this end, a unified DOB applicable to both flight and land modes is proposed. 

First, the vehicle dynamics in both modes are abstracted into a unified second-order system:
\begin{equation}
    \ddot{\mathbf{r}}=\mathbf{u}+\mathbf{d}
    \label{dob:1}
\end{equation}
where $\ddot{\mathbf{r}}=[\ddot{x},\ddot{y},\ddot{z}]^T\in \mathbb{R}^3$ is the acceleration of the bimodal vehicle, $\mathbf{u}$ and $\mathbf{d}$ denote the control input and disturbance vectors, respectively, which can be derived from the air–land bimodal dynamics model (\ref{eq:trans}), (\ref{eq:ground-dynamic-uni}). Specifically, $\mathbf{u} \in \{\mathbf{u}_1, \mathbf{u}_2\}$ and $\mathbf{d} \in \{\mathbf{d}_1, \mathbf{d}_2\}$, depending on the current locomotion mode. When operating in quadrotor flight mode:
\begin{equation}
    \begin{aligned}
        \mathbf{u}_1=F_1{\mathbf{R}}_3(\mathbf{\Theta})/m ,\\
        \mathbf{d}_1=-\mathbf{g}-\mathbf{d}_\text{air}/{m},
    \end{aligned}
    \label{dob:air-ud}
\end{equation}
when operating in wheeled land mode:
\begin{equation}
    \begin{aligned}
        \mathbf{u}_2=\mathbf{M}^{-1}\mathbf{C}(\psi)\mathbf{F} ,\\
        \mathbf{d}_2=-\mathbf{M}^{-1}\mathbf{H}(\psi)\mathbf{d}_\text{land}.
    \end{aligned}
    \label{dob:land-ud}
\end{equation}

Second, we formulate a real-time disturbance estimation method based on the Uncertainty and Disturbance Estimator (UDE) proposed by Zhang et al. \cite{zhang2018ude}. In this approach, a first-order low-pass filter 
\begin{equation}
    G(s) = \frac{1}{Ts+1}
\end{equation}
is employed to separate low-frequency disturbances from high-frequency measurement noise, where $T$ denotes the filter time constant. Defining 
\[
    \mathcal{F}_T\{\cdot\} \;\triangleq\; \text{the output of the filter } G(s),
\]
the disturbance estimate in the time domain can be written as
\begin{equation}
    \hat{\mathbf{d}}(t) = \mathcal{F}_T\!\left\{\,\ddot{\mathbf{r}}(t) - \mathbf{u}(t)\,\right\},
    \label{dob:10-t}
\end{equation}
where $\ddot{\mathbf{r}}(t)$ is the IMU-measured acceleration.


Finally, this unified formulation enables disturbances in both flight and land modes to be estimated through a common observer structure, thereby simplifying implementation and ensuring consistent disturbance-aware planning and control across modes.

\subsection{Safety Boundary Adjustment}
\label{sect:Safety Boundary Adjustment}
Physically, the control inputs $\mathbf{u}_1$ and $\mathbf{u}_2$ are themselves bounded due to the physical actuation limits of the system: in flight mode, the total thrust is upper-bounded by the maximum rotor force, while in land mode the wheel actuation is bounded by the maximum driving torque. 
In both modes, the control input is bounded by the physical actuation limits of the vehicle. Specifically,
\begin{equation}
    \mathbf{u} =
    \begin{cases}
        \mathbf{u}_1 = \dfrac{1}{m}\mathbf{R}_3(\mathbf{\Theta})F_1, & 
        0 \leq F_1 \leq F_{1,\max}, \\[10pt]
        \mathbf{u}_2 = \mathbf{M}^{-1}\mathbf{C}(\psi)\mathbf{F}, &
        \|\mathbf{F}\| \leq F_{2,\max},
    \end{cases}
    \label{eq:u1u2-def}
\end{equation}
where $\mathbf{u}_1$ corresponds to the flight mode with rotor thrust $F_1$, and $\mathbf{u}_2$ corresponds to the land mode with wheel driving force vector $\mathbf{F}$.  
The boundedness of $\mathbf{u}$ in the world frame can therefore be written axis-wise as
\begin{equation}
    -\frac{F_{i,\max}^{\mu}}{m} \;\leq\; u_{i}^{\mu} \;\leq\; \frac{F_{i,\max}^{\mu}}{m}, 
    \quad \mu \in \{x_w, y_w, z_w\}, \; i \in \{1,2\},
    \label{eq:u-bound}
\end{equation}
where $F_{i,\max}^{\mu}$ denotes the maximum achievable force along the $\mu$-axis in mode $i$, after transforming the body-frame actuation limits into the world frame.  
Substituting \eqref{eq:u-bound} together with the disturbance estimates $\hat{\mathbf{d}}_1$ and $\hat{\mathbf{d}}_2$ into \eqref{dob:1} leads directly to the acceleration boundaries. In flight mode:
\begin{equation}
    \left\{
    \begin{array}{l}
    \left[a_{1,\min}^{x_w}, a_{1,\max}^{x_w}\right] = 
    \left[\frac{{ - F_{1,\max}^{x_w}}}{m}+\hat{d}_{1,1},\frac{{F_{1,\max}^{x_w}}}{m}+\hat{d}_{1,1}\right] \\[8pt]
    \left[a_{1,\min}^{y_w}, a_{1,\max}^{y_w}\right] = 
    \left[\frac{{ - F_{1,\max}^{y_w}}}{m}+\hat{d}_{1,2},\frac{{F_{1,\max}^{y_w}}}{m}+\hat{d}_{1,2}\right] \\[8pt]
    \left[a_{1,\min}^{z_w}, a_{1,\max}^{z_w}\right] = 
    \left[\hat{d}_{1,3},\frac{{F_{1,\max}^{z_w}}}{m}+\hat{d}_{1,3}\right]
    \end{array},
    \right.
\label{eq:acceleration_limitation_air}
\end{equation}
and in land mode:
\begin{equation}
    \left\{
    \begin{array}{l}
    \left[a_{2,\min}^{x_w}, a_{2,\max}^{x_w}\right] = 
    \left[\frac{{ - F_{2,\max}^{x_w}}}{m}+\hat{d}_{2,1},\frac{{F_{2,\max}^{x_w}}}{m}+\hat{d}_{2,1}\right] \\[8pt]
    \left[a_{2,\min}^{y_w}, a_{2,\max}^{y_w}\right] = 
    \left[\frac{{ - F_{2,\max}^{y_w}}}{m}+\hat{d}_{2,2},\frac{{F_{2,\max}^{y_w}}}{m}+\hat{d}_{2,2}\right] \\[8pt]
    \left[a_{2,\min}^{z_w}, a_{2,\max}^{z_w}\right] = 0
    \end{array}
    \right.
\label{eq:acceleration_limitation_land}
\end{equation}
where $a_{1,\min}^{\mu}$ and $a_{1,\max}^{\mu}$ ($\mu \in \{x_w, y_w, z_w\}$) denote the axis-wise acceleration limits in the quadrotor flight mode, while $a_{2,\min}^{\mu}$ and $a_{2,\max}^{\mu}$ denote the corresponding limits in the wheeled land mode. 

\section{Constraint-Adaptive Bimodal Motion Planning}

In this section, we present a constraint-adaptive bimodal motion planner that operates within the dynamically adjusted feasible constraint boundaries. The planner consists of two integrated components: 1) a disturbance-aware kinodynamic path-searching module that directs the initial trajectory search toward regions with reduced disturbance, and 2) a B-spline-based trajectory optimization module that refines the trajectory while enforcing dynamic feasibility within time-varying constraint boundaries.

\subsection{Disturbance-Aware Kinodynamic Path Searching}

Path searching serves as the front end of the proposed planner, aiming to generate an initial feasible trajectory that connects the current state to the target state while satisfying the bimodal vehicle's dynamic constraints. We employ the kinodynamic A* framework \cite{zhou2019robust}, which uses motion primitives generated by integrating discrete acceleration inputs ($\mathbf{a}=[a^{x_w},a^{y_w},a^{z_w}]\in\mathbb{R}^3$) over time to connect the current state with the target state. Within this framework, the trajectory cost is defined as 
\begin{equation}
    f_c = g_c + h_c, 
\end{equation}
where $g_c$ represents the cumulative motion primitive cost and $h_c$ denotes the heuristic cost estimating the remaining effort required to reach the goal.

Building upon this foundation, we introduce disturbance awareness into the heuristic function to enhance planning robustness in disturbed environments. Specifically, two extensions are proposed: 1) an altitude-based energy cost $F_e(z_m)$ to discourage aerial trajectories and encourage energy-efficient ground motion, and 2) a disturbance-aware directional penalty $F_d(\mathbf{a})$ to guide the search toward directions with greater dynamic feasibility. 

The altitude-based energy cost is defined as
\begin{equation}
    F_e(z_m) = 
    \begin{cases} 
        (z_m - z_{{thr}})^2, & z_m > z_{{thr}} \\
        0, & z_m \leq z_{{thr}}
    \end{cases}
    \label{eq:energy cost}
\end{equation}
where $z_m$ denotes the altitude of the endpoint of the motion primitive and $z_{{thr}}$ specifies the threshold distinguishing ground from aerial motion primitives. 

To further enhance robustness, we define the disturbance-aware directional penalty as
\begin{equation}
    F_d(\mathbf{a}) = \frac{1}{\epsilon+\text{c}(\mathbf{a})}
    \label{eq:search direction cost}
\end{equation}
where $\epsilon=10^{-3}$ is a small constant to prevent numerical singularities. The function $\text{c}(\mathbf{a})$ measures the distance of the acceleration input from the feasible dynamic boundaries:
\begin{equation}
\text{c}(\mathbf{a}) = \left\{
\begin{aligned}
    & \quad\sum_{{\mu\in\{x_w,y_w,z_w\}}} \min\bigl( a^{\mu} - a_{1,\min}^{\mu}, a_{1,\max}^{\mu} - a^{\mu} \bigr), \\
    & \qquad\qquad z_m > z_{thr}; \\
    & \quad\sum_{{\mu}\in\{x_w,y_w\}} \min\bigl( a^{\mu} - a_{2,\min}^{\mu}, a_{2,\max}^{\mu} - a^{\mu} \bigr), \\
    & \qquad\qquad z_m \leq z_{thr}
\end{aligned}
\right.
\end{equation}
where $a_{1,\min}^\mu$, $a_{1,\max}^\mu$, $a_{2,\min}^\mu$, and $a_{2,\max}^\mu$ represent axis-wise acceleration bounds for the air and land modes, respectively. These bounds are adaptively adjusted by (\ref{eq:acceleration_limitation_air}), (\ref{eq:acceleration_limitation_land}).

The overall cost function becomes 
\begin{equation}
    f_c = g_c + h_c +  F_e(z_m) + F_d(\mathbf{a}).
\end{equation}

Based on this cost function, path searching proceeds as follows. First, from each node, a set of candidate motion primitives is generated by integrating discrete acceleration inputs $\mathbf{a}$ over a fixed time step, thereby predicting the vehicle’s future states. Each primitive is evaluated using the overall cost function: $g_c$ accumulates the execution cost of the primitive (e.g., time or control effort), $h_c$ provides a heuristic estimate of the distance-to-goal cost, $F_e(z_m)$ imposes altitude penalties to discourage flight when ground motion is feasible, and $F_d(\mathbf{a})$ ensures that primitives close to the dynamic boundaries are disfavored. Next, infeasible primitives that violate obstacle constraints or dynamic limits are pruned. Among the remaining candidates, the primitive with the lowest cost is expanded, and the process repeats until the goal region is reached. Finally, the trajectory with the minimum cost $f_c$ is selected and provided as the input to the B-spline trajectory optimization.

\subsection{B-spline Trajectory Optimization with Time-Varying Dynamic Boundaries}

The disturbance-aware path-searching module provides an initial feasible trajectory. However, this trajectory often lacks smoothness and strict adherence to the bimodal vehicle’s dynamic constraints. To address these issues, we refine the trajectory using B-spline optimization. The initial trajectory is parameterized as a $p_b$-degree B-spline with $N+1$ control points $\mathbf{Q}=\{\mathbf{Q}_0,\mathbf{Q}_1,...,\mathbf{Q}_i,...,\mathbf{Q}_N\}$, where $\mathbf{Q}_i=[Q_i^x,Q_i^y,Q_i^z]\in\mathbb{R}^3$ represents the position of $i$-th control point. The terrestrial control points $\mathbf{Q}_t=\{\mathbf{Q}_{t0},\mathbf{Q}_{t1},...,\mathbf{Q}_{tM}\}$ are identified by the height threshold $Q_i^{z}\le z_{{thr}}$. Owing to the convex hull property of B-splines, constraining the control points with respect to smoothness, obstacle avoidance, dynamic feasibility, and curvature ensures that the resulting trajectory satisfies these requirements.

Following Zhang et al. \cite{zhang2022autonomous}, we first adopt the control-point cost formulation:
\begin{equation}
    f_{total} = \lambda_s f_s + \lambda_l f_l + \lambda_f( f_v + f_a) + \lambda_n f_n
    \label{eq:opt_cost_function}
\end{equation}
where $\lambda_s$, $\lambda_l$, $\lambda_f$, and $\lambda_n$ are the corresponding weights for each cost term. 

The smoothness cost $f_s$ is defined as
\begin{equation}
    f_s=\sum_{i=p_b-1}^{N-p_b+1}\|\left(\mathbf{Q}_{i+1}-\mathbf{Q}_i\right)+\left(\mathbf{Q}_{i-1}-\mathbf{Q}_i\right)\|^2
\end{equation}
which penalizes large variations between neighboring control points, thereby promoting high-order continuity of the trajectory and ensuring smooth velocity and acceleration profiles.

The collision cost $f_l$ is formulated as
\begin{equation}
f_l=\sum_{i=p_b}^{N-p_b} F_l\left(d\left(\mathbf{Q}_i\right)\right),
\end{equation}
\begin{equation}
    F_l\left(d\left(\mathbf{Q}_i\right)\right)=\left\{\begin{array}{cl}\left(d\left(\mathbf{Q}_i\right)-d_{t h r}\right)^2 & d\left(\mathbf{Q}_i\right) \leq d_{t h r} \\ 0 & d\left(\mathbf{Q}_i\right)>d_{t h r}\end{array}\right.,
\end{equation}
$d(\mathbf{Q}_i)$ is the distance from the control point to the nearest obstacle and $d_{thr}$ is the minimum safety distance. This cost penalizes control points within the safety margin, thereby enforcing collision-free trajectory generation.

The velocity cost $f_v$ is formulated as
\begin{equation}
    f_v=\sum_{\substack{\mu \in \\\{x_w, y_w, z_w\}}} \sum_{i=p_b-1}^{N-p_b} F_v\left(v_{i}^{\mu}\right),
\end{equation}
\begin{equation}
    F_v\left(v_{i}^{\mu}\right)=\left\{\begin{array}{cc}
\left(\left({v_{i}^{\mu}}\right)^2-v_{max }^2\right)^2 & \left({v_{i}^{\mu}}\right)^2>v_{max }^2 \\
0 & \left({v_{i}^{\mu}}\right)^2 \leq v_{max }^2
\end{array}\right.
\end{equation}
where $v_{i}^{\mu}$ is the velocity of the $i$-th control point along axis $\mu$ and $v_{max}$ is the maximum allowable velocity. This cost penalizes velocity violations.

The curvature cost $f_n$ is defined as
\begin{equation}
    f_n=\sum_{i=1}^{M-1} F_n\left(\mathbf{Q}_{t i}\right),
\end{equation}
\begin{equation}
    F_n\left(\mathbf{Q}_{t i}\right)= \begin{cases}\left(\mathbf{C}_i-\mathbf{C}_{max }\right)^2 & \mathbf{C}_i>\mathbf{C}_{max } \\ 0 & \mathbf{C}_i \leq \mathbf{C}_{max }\end{cases}
\end{equation}
where $\mathbf{C}_i$ is the curvature at the terrestrial control point $\mathbf{Q}_{t i}$, $\mathbf{C}_{max}$ is the curvature threshold. This cost limits the curvature of the terrestrial trajectory.

To incorporate disturbance adaptation, the trajectory optimization is constrained within dynamically adjusted boundaries (\ref{eq:acceleration_limitation_air}), (\ref{eq:acceleration_limitation_land}). Acceleration feasibility is enforced through a smooth penalty term defined by the softplus function:
\begin{equation}
f_a = \sum_{\mu \in \{x_w,y_w,z_w\}} \sum_{i=p_b - 2}^{N - p_b} F_a(a_i^\mu),
\label{eq:acc constraint}
\end{equation}
$a_i^{\mu}$ is the acceleration of the $i$-th control point along axis $\mu$,
\begin{equation}
    F_a(a_i^\mu) =
\left( \text{softplus}\left(a_{\min}^{\mu} - a_i^\mu \right) \right)^2 +
\left( \text{softplus}\left(a_i^\mu - a_{\max}^{\mu} \right) \right)^2,
\end{equation}
\begin{equation}
\text{softplus}(x) = \frac{1}{\beta} \log\left(1 + e^{\beta x} \right),
\end{equation}
$\beta$ is the smoothness parameter of the softplus function. The time-varying acceleration bounds \(a_{\min}^{\mu}, a_{\max}^{\mu}\) are determined according to the control point height:
\begin{equation}
a_{\min}^{\mu}, a_{\max}^{\mu} =
\begin{cases}
a_{1,\min}^{\mu},\ a_{1,\max}^{\mu}, & \text{if } Q_i^z > z_{thr}\\
a_{2,\min}^{\mu},\ a_{2,\max}^{\mu}, & \text{if } Q_i^z \leq z_{thr}
\end{cases}.
\end{equation}

The optimization process is expressed as follows. First, the initial trajectory obtained from path searching is expressed in B-spline form. The control points are then iteratively adjusted to minimize the total cost in (\ref{eq:opt_cost_function}).


This nonlinear optimization problem is solved using the NLopt library \footnote{\url{https://github.com/stevengj/nlopt/releases}}. After trajectory optimization is completed, the resulting trajectory is passed to the controller for execution. The motion modality is determined according to the control point height: segments with $Q_i^z>z_{thr}$ are classified as aerial trajectories and executed by the flight controller, whereas segments with $Q_i^z \leq z_{thr}$ are identified as terrestrial trajectories and executed by the land-mode controller.

\section{Results}

\subsection{Robot Platform}

\begin{figure}[ht]
    \centering
    \includegraphics[width=\linewidth]{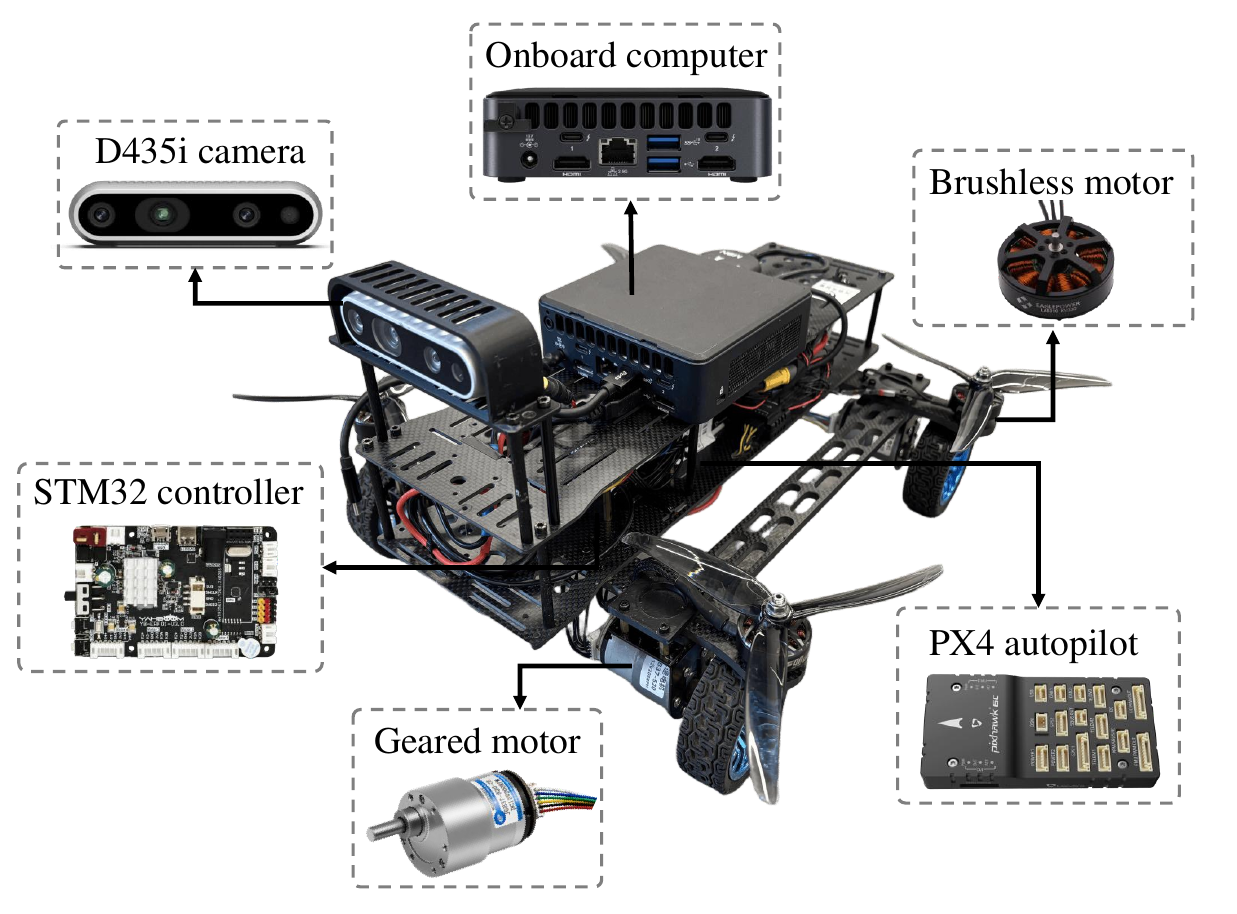}
    \caption{Hardware structure of the self-developed air–land bimodal vehicle}
    \label{fig:platform}
\end{figure}

The self-developed air–land bimodal vehicle (Fig.~\ref{fig:platform}) integrates two power systems: four brushless motors with propellers for flight, and four geared motors with wheels for ground mobility. A RealSense D435i depth camera provides depth perception for local mapping, while an Intel NUC11TNKi5 onboard computer executes the navigation framework. A Pixhawk 6C handles flight control, and an STM32 board controls ground locomotion. All hardware components are mounted on a 3D-printed carbon fiber frame powered by a 19V 5S lithium battery.

\subsection{Simulations}

\begin{figure}[htbp]
\centering
\includegraphics[width=\linewidth]{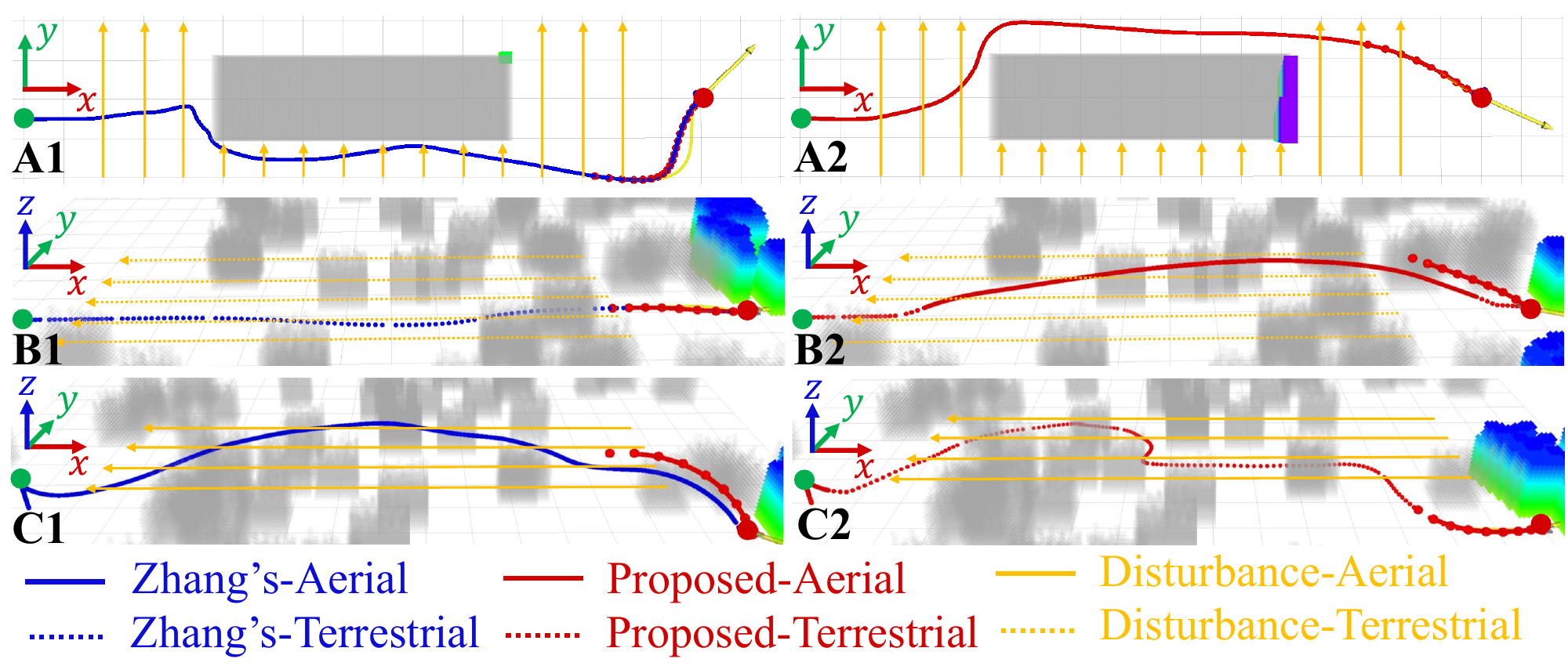}
\caption{Comparisons of terrestrial–aerial trajectories from simulations. Subfigures A–C illustrate the trajectory results for the Urban Crosswind, High-Resistance Terrain, and Extreme Wind scenarios, respectively.}
\label{fig:trajs-sim}
\end{figure}

To assess the performance of the proposed framework, we compare it with a representative air-land bimodal navigation framework \cite{zhang2022autonomous} across three simulated disturbance scenarios. Fig.~\ref{fig:trajs-sim} shows the trajectories, while TABLE~\ref{tab:benchmark} summarizes the quantitative performance metrics. In the simulations, the control inputs are the rotational speeds of the four motors. The mechanical power of each motor is calculated as $P = \tau \omega$, where the torque is modeled as $\tau = k_{\text{torque}} \cdot \text{rpm}^2$ and the angular velocity is given by $\omega = \frac{2\pi}{60} \cdot \text{rpm}$. The total system energy consumption is obtained by integrating the sum of the four motors’ mechanical power over time.

\subsubsection{Urban Crosswind Scenario}

This simulation aims to evaluate the planner’s adaptability to crosswind disturbances in an urban environment, where obstacles are modeled as buildings and lateral winds are commonly encountered. As summarized in TABLE~\ref{tab:benchmark}, the proposed method achieved a 30\% reduction in task completion time, a 30\% decrease in energy consumption, and a 29\% improvement in tracking accuracy compared with the baseline. These improvements are primarily attributed to the disturbance-aware search-direction heuristic in (\ref{eq:search direction cost}), which guided the trajectory toward the positive $y$-axis as shown in Fig.~\ref{fig:trajs-sim}A2. By leveraging building shielding to mitigate crosswind effects, the planner maintained forward acceleration and significantly enhanced navigation efficiency and accuracy.

\begin{table}[htbp]
\centering
\caption{Performance Comparison in Simulation}
\resizebox{\linewidth}{!}{\begin{tabular}{ccccc}
\toprule
Scenario & Method & Time (s) & Energy (Wh) & RMSE (m) \\
\midrule
\multirow{2}{*}{Urban Crosswind}
& Zhang’s \cite{zhang2022autonomous} & 13.541 & 0.550 & 0.2781 \\
& Proposed & \textbf{9.542} & \textbf{0.385} & \textbf{0.1968} \\
\midrule
\multirow{2}{*}{High-Resistance Terrain}
& Zhang’s \cite{zhang2022autonomous} & 11.454 & \textbf{0.466} & 0.3032 \\
& Proposed & \textbf{9.384} & 0.507 & \textbf{0.2625} \\
\midrule
\multirow{2}
{*}{Extreme Wind}
& Zhang’s \cite{zhang2022autonomous} & 11.778 & 0.705 & 0.3628 \\
& Proposed & \textbf{10.717} & \textbf{0.610} & \textbf{0.3479} \\
\bottomrule
\end{tabular}}
\label{tab:benchmark}
\end{table}

\subsubsection{High-Resistance Terrain Scenario}

This simulation evaluates the planner’s adaptability to strong ground disturbances caused by rugged terrain. As shown in TABLE~\ref{tab:benchmark}, the proposed method reduced task duration by 18\% and improved tracking accuracy by 13\%, with only a 9\% increase in energy consumption compared with the baseline. These improvements stem from the disturbance-adaptive safety-boundary adjustment mechanism, which detected excessive ground disturbances and dynamically contracted the feasible dynamic boundaries for the land mode (\ref{eq:acceleration_limitation_land}). Consequently, the disturbance-aware path-searching module was guided to select a lower-cost aerial trajectory over the infeasible ground alternative (Fig.~\ref{fig:trajs-sim}B2). By avoiding inefficient ground motion under severe disturbances, the proposed framework achieved more robust and reliable navigation performance.

\subsubsection{Extreme Wind Scenario}

This simulation evaluates the planner’s adaptability to intense aerial disturbances induced by extreme wind conditions. As shown in TABLE~\ref{tab:benchmark}, the proposed method decreased task duration by 9\%, enhanced tracking accuracy by 4\%, and lowered energy consumption by 13\% relative to the baseline. In this simulation, the baseline method ignored aerial disturbances and generated an aerial trajectory (Fig.~\ref{fig:trajs-sim}C1), which was substantially disrupted by strong headwinds. In contrast, the disturbance-adaptive safety-boundary adjustment mechanism detected severe aerial disturbances and dynamically contracted the feasible dynamic boundaries of the flight mode (\ref{eq:acceleration_limitation_air}). As a result, the planner generates a lower-cost ground trajectory (Fig.~\ref{fig:trajs-sim}C2). By adaptively favoring terrestrial navigation under strong winds, the planner effectively mitigated aerodynamic effects and achieved more robust overall performance.

\subsection{Experiments}

\begin{figure}[htbp]
\centering
\includegraphics[width=\linewidth]{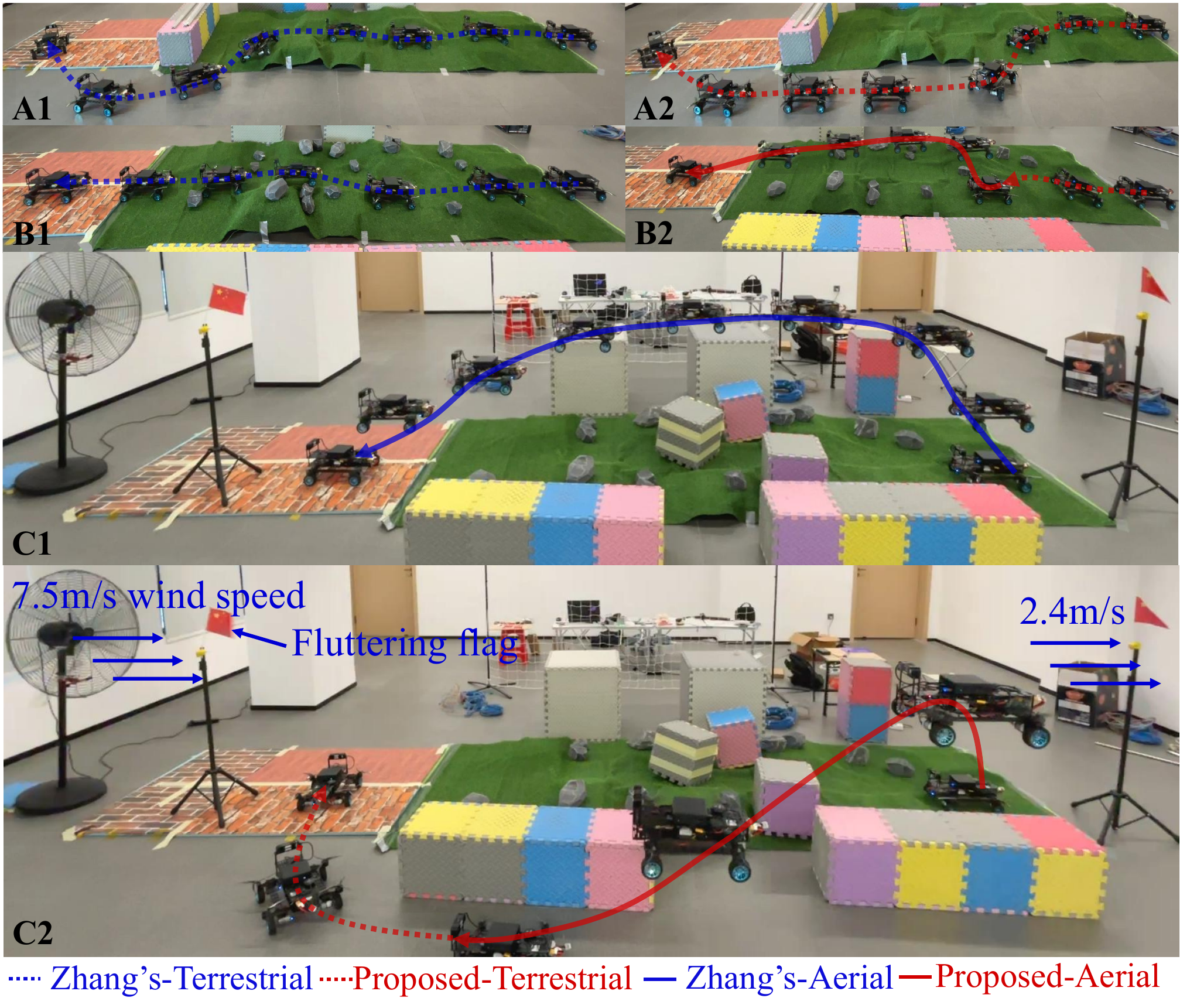}
\caption{Comparisons of terrestrial–aerial trajectories obtained from real-world experiments. Subfigures A–C illustrate the corresponding experimental trajectories for the Rugged Terrain Navigation, Blocked Terrain Navigation, and Headwind Environment scenarios.}
\label{fig:trajs-exp}
\end{figure}

To validate the proposed framework in real-world environments, two types of disturbances—terrestrial and aerial—were configured, as illustrated in Fig.~\ref{fig:airflow}, and three sets of experiments were conducted accordingly. The corresponding trajectories and quantitative results are summarized in Fig.~\ref{fig:trajs-exp} and TABLE~\ref{tab:experiments}, while Fig.~\ref{fig:ude} presents the estimated disturbances during execution. In experiments, each planning cycle—including both path searching and trajectory optimization—was completed within approximately 10~ms on the self-developed platform (Fig.~\ref{fig:platform}). This computational efficiency enables high-frequency planning, thereby ensuring that the proposed framework can effectively operate in dynamic environments with real-time disturbances.

\subsubsection{Rugged Terrain Navigation}

This experiment evaluates the planner’s adaptability to ground disturbances in unstructured terrain. In this scenario, the flight-energy heuristic in (\ref{eq:energy cost}) encouraged ground locomotion, while the search-direction heuristic in (\ref{eq:search direction cost}) and the dynamic feasibility cost in (\ref{eq:acc constraint}) guided the trajectory search toward dynamically feasible and less disturbed regions. As a result, the proposed planner generated a leftward detour trajectory around the first major obstacle (Fig.~\ref{fig:trajs-exp}A2), effectively avoiding the disturbed area. Compared with the baseline \cite{zhang2022autonomous}, the proposed method reduced task duration by 17\%, energy consumption by 18\%, and tracking error by 29\% (TABLE~\ref{tab:experiments}). These improvements highlight the effectiveness of incorporating disturbance-aware heuristics in maintaining efficiency and tracking accuracy under challenging ground conditions.

\begin{table}[htbp]
\centering
\caption{Performance Comparison In Real World Experiments}
\label{tab:experiments}
\resizebox{\linewidth}{!}{\begin{tabular}{ccccc}
\toprule
Scenario & Method & Time (s) & Energy (Wh) & RMSE (m) \\
\midrule
\multirow{2}{*}{Rugged Terrain}
& Zhang’s \cite{zhang2022autonomous} & 30.13 & 0.690 & 0.1599 \\
& Proposed & \textbf{25.10} & \textbf{0.563} & \textbf{0.1142} \\
\midrule
\multirow{2}{*}{Blocked Terrain}
& Zhang’s \cite{zhang2022autonomous} & 27.37 & \textbf{0.624} & 0.1777 \\
& Proposed & \textbf{22.93} & 1.864 & \textbf{0.1544} \\
\midrule
\multirow{2}
{*}{Headwind}
& Zhang’s \cite{zhang2022autonomous} & \textbf{27.53} & 4.366 & 0.3402 \\
& Proposed & 29.07 & \textbf{2.320} & \textbf{0.1362} \\
\bottomrule
\end{tabular}}
\end{table}

\subsubsection{Blocked Terrain Navigation}

\begin{figure}[htbp]
\centering
\includegraphics[width=\linewidth]{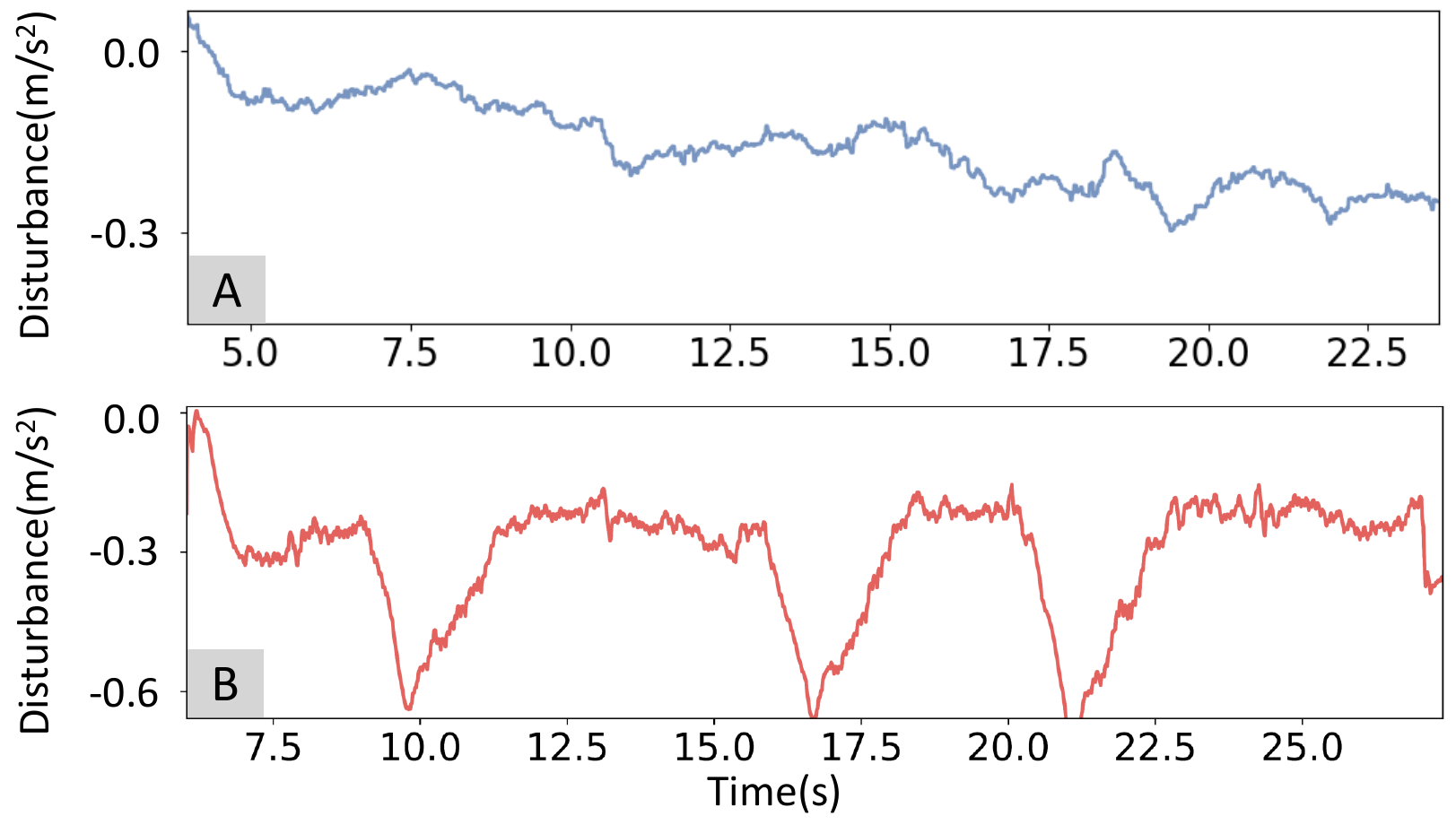}
\caption{Estimated $x$ axis disturbance in experiments. (A) corresponds to Experiment “Headwind Environment” (Fig. \ref{fig:trajs-exp}C1). (B) corresponds to Experiment “Blocked Terrain Navigation” (Fig. \ref{fig:trajs-exp}B1).}
\label{fig:ude}
\end{figure}

This experiment evaluates the planner’s decision-making capability when ground navigation becomes infeasible due to severe disturbances and blocked detour paths. Based on the rugged terrain setup, with additional obstacles placed on both sides of the first major obstacle to eliminate potential ground detours. Upon detecting strong ground disturbances and no feasible terrestrial trajectory, the proposed method planned an aerial trajectory (Fig.~\ref{fig:trajs-exp}B2). Compared with the baseline, it shortened task duration by 16\% and reduced tracking error by 13\%, albeit with increased energy consumption due to aerial locomotion (TABLE~\ref{tab:experiments}). These results demonstrate that the disturbance-aware motion-planning framework effectively balances feasibility and performance in complex ground conditions. The estimated disturbance profile in Fig.~\ref{fig:ude}B further corroborates this finding, as the noticeable valleys correspond to uneven terrain segments where ground disturbances hindered feasible navigation, leading the planner to generate an aerial trajectory.

\subsubsection{Headwind Environment}

This experiment assesses the planner’s adaptability to strong aerial disturbances. In an environment with strong headwinds generated by an industrial fan, the proposed method planned a leftward detour trajectory after takeoff and transitioned to land mode to complete the task (Fig.~\ref{fig:trajs-exp}C2). Compared with the baseline, it reduced energy consumption by 47\% and tracking error by 60\%, though with a 6\% increase in task duration (TABLE~\ref{tab:experiments}). These results demonstrate the framework’s ability to trade off between stability and efficiency under extreme aerial disturbances. As shown in Fig.~\ref{fig:ude}A, the estimated disturbance magnitude increases significantly as the vehicle approaches the headwind source, validating the need for switching to ground mode to maintain safety and performance.

\begin{figure}[htbp]
\centering
\includegraphics[width=\linewidth]{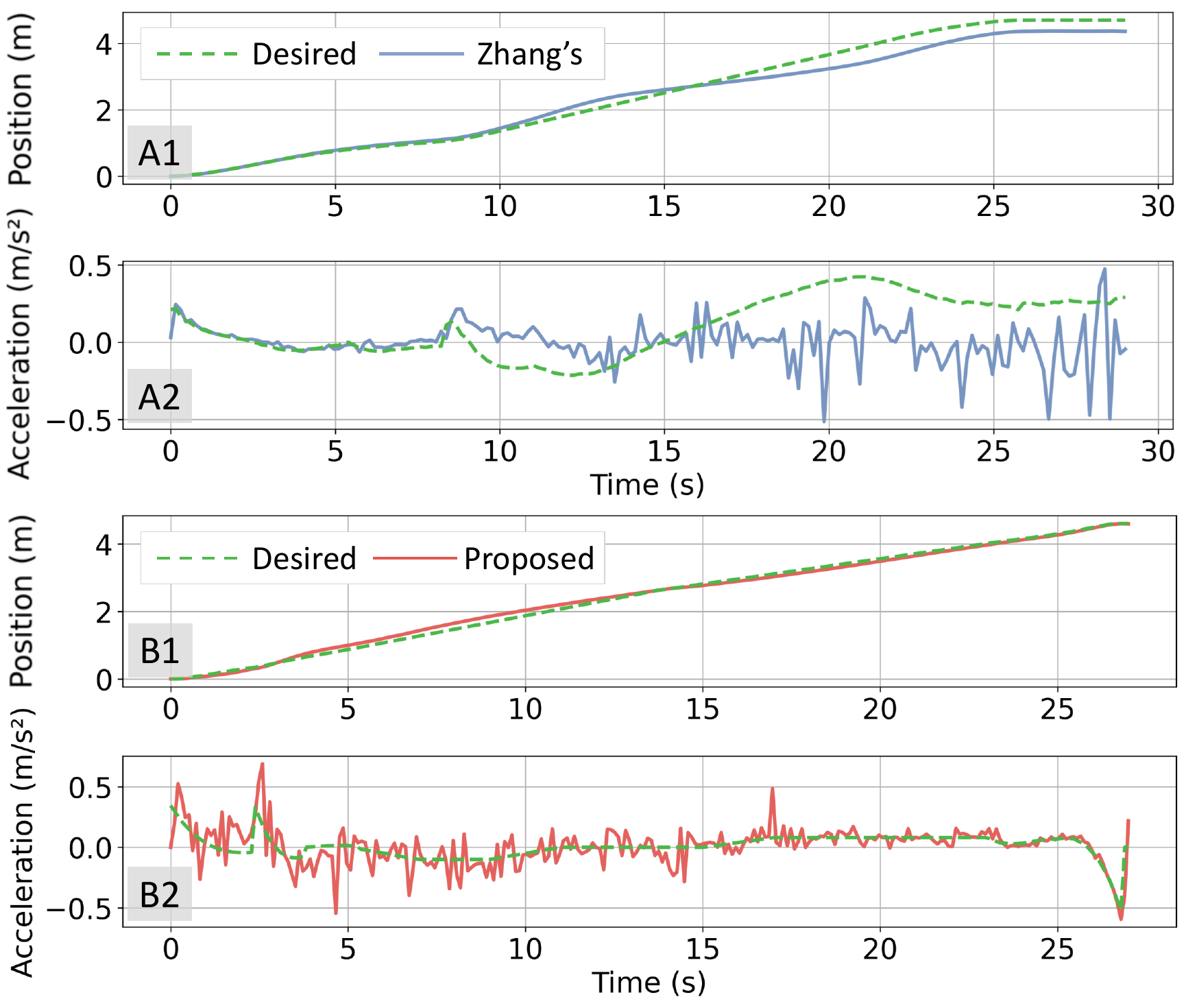}
\caption{Tracking results of the $x$ axis states in the Headwind experiment. (A) Baseline method (B) Proposed method.}
\label{fig:tracking-data}
\end{figure}

Fig.~\ref{fig:tracking-data} shows the tracking performance of the proposed method compared with Zhang’s method \cite{zhang2022autonomous}. To ensure a fair comparison, the same PID controller is employed for both. The baseline method does not account for aerial disturbances and generates an aerial trajectory that is significantly disrupted by strong winds. The resulting aerodynamic load leads to actuator saturation, preventing the system from following the desired acceleration  (Fig.~\ref{fig:tracking-data}A2), and ultimately causes large position errors (Fig.~\ref{fig:tracking-data}A1). In contrast, the proposed planner detects the strong headwind and generates a ground-based detour trajectory, preserving dynamic feasibility and enabling accurate trajectory tracking (Fig.~\ref{fig:tracking-data}B).

\section{Conclusions}

This paper proposes a disturbance-aware motion-planning framework for air–land bimodal vehicles to address the challenges of bimodal dynamics and unknown disturbances. By integrating a unified disturbance observer, adaptive constraint adjustment, kinodynamic path searching, and B-spline trajectory optimization, the framework enables robust and efficient trajectory generation in disturbed environments. Simulations and real-world experiments demonstrate its effectiveness, showing reduced trajectory-tracking error and improved time–energy trade-offs relative to a representative baseline \cite{zhang2022autonomous}. These results highlight the potential of disturbance-aware planning to enhance the robustness and efficiency of air–land bimodal navigation. Future work will extend the framework to multi-robot systems and more complex environments.
\bibliography{Bibtex}

\addtolength{\textheight}{-12cm}   
                                  
\end{document}